\pdfoutput=1
\documentclass[11pt]{article}

\usepackage[]{acl}

\usepackage{times}
\usepackage{latexsym}

\usepackage[T1]{fontenc}
\usepackage{etoolbox}

\usepackage[utf8]{inputenc}
\usepackage{adjustbox}
\usepackage{microtype}

\usepackage{booktabs, multirow}
\usepackage{subfiles}

\usepackage{graphicx}

\title{MasonNLP+ at SemEval-2023 Task 8: Extracting Medical Questions, Experiences and Claims from Social Media using Knowledge-Augmented Pre-trained Language Models}

\newcommand{\authinfo}[2]{\fontsize{11pt}{12pt}\selectfont #1\textsuperscript{#2}}
\newcommand{\authinfofoot}[3]{\fontsize{11pt}{12pt}\selectfont #1\textsuperscript{#2}#3}

\author{
\authinfofoot{Giridhar Kaushik Ramachandran}{1}{\thanks{\quad Equal contribution}}, \authinfo{Haritha Gangavarapu}{2}{\footnotemark[1]}, 
\authinfo{Kevin Lybarger}{1}, 
\authinfo{\"{O}zlem Uzuner}{1} \\
\textsuperscript{1}Department of Information Sciences and Technology, George Mason University \\
\textsuperscript{2} CGI Inc.
}

\begin{document}

\maketitle

\newif\ifsubfile
\subfiletrue

\begin{abstract}

In online forums like Reddit, users share their experiences with medical conditions and treatments, including making claims, asking questions, and discussing the effects of treatments on their health. Building systems to understand this information can effectively monitor the spread of misinformation and verify user claims. The Task-8 of the 2023 International Workshop on Semantic Evaluation focused on medical applications, specifically extracting patient experience- and medical condition-related entities from user posts on social media. The Reddit Health Online Talk (RedHot) corpus contains posts from medical condition-related subreddits with annotations characterizing the patient experience and medical conditions. In Subtask-1, patient experience is characterized by personal experience, questions, and claims. In Subtask-2, medical conditions are characterized by population, intervention, and outcome. For the automatic extraction of patient experiences and medical condition information, as a part of the challenge, we proposed language-model-based extraction systems that ranked $3^{rd}$ on both subtasks' leaderboards. In this work, we describe our approach and, in addition, explore the automatic extraction of this information using domain-specific language models and the inclusion of external knowledge.

\end{abstract}

\section{Introduction}

Social media platforms, like Reddit, allow users to share their experiences pseudonymously.  On these platforms, users tend to share different kinds of information, including questions, their personal experiences using certain prescribed drugs, or how they deal with the different symptoms caused by their medical condition. Users may also make claims about certain clinical and non-clinical interventions (e.g. home remedies) and the associated outcomes. While this information can provide insight into medical conditions and how the condition presentation among users, their personal experience may be inconsistent with current medical evidence.

Thus, effective monitoring of these medical condition-related forums is needed to prevent the spread of misinformation and corroborate various user claims against scientific evidence. Information from social media could also augment current scientific understanding of conditions and treatments. Natural language processing (NLP) information extraction (IE) techniques can extract information from user posts, which could be incorporated into systems for monitoring and preventing user misconception. The RedHot \citep{redhot2022} corpus consists of over 22,000 user posts from 24 medical condition-related subreddits annotated for entities related to patient experiences, claims, interventions, and outcomes. The posts are annotated in two stages: (1) Subtask-1 annotations include user \textit{claims}, \textit{personal experiences}, \textit{questions}, and \textit{claim-based personal experiences} and (2) Subtask-2 annotations contain more granular information through entities related to patient \textit{population}, \textit{intervention}, and \textit{outcome} (PIO). 

In this work, we explore the automatic extraction of these entities from a subset of the RedHot corpus as released through Task-8\footnote{\url{https://causalclaims.github.io/}} \cite{khetan-EtAl:2023:SemEval} of the 17th workshop on Semantic Evaluation. The extraction of the patient experience and PIO entities was pursued in Subtask-1 and Subtask-2, respectively. We propose Bidirectional Encoder Representations from Transformer (BERT)-based systems that extract the patient-experience and medical-condition entities. In this work, we compare BERT-based systems pre-trained on the general domain with domain-adapted BERT models and explore augmentation techniques incorporating external clinical knowledge priors. We extract the patient experience entities from Subtask-1 and medical condition entities from Subtask-2 at an overall F1 of 68.59 (sentence-level) and 32.65 (token-level), respectively, on the challenge test set. Our BERT-based systems ranked 3\textsuperscript{rd} in both subtasks of the challenge. 

\section{Related Work}

Extracting medical information from free text is a well-explored research domain. This medical IE work can broadly be viewed as belonging to two major categories: (1) Work attempting to extract information from biomedical literature and clinical text from electronic health records and (2) work that focuses on extracting relevant medical information from public forums, user platforms, and social media. 
Medical IE work includes the development of annotated data sets and data-driven extraction architectures. We discuss relevant datasets first and then briefly review extraction approaches.

In biomedical literature, commonly extracted information includes medical problems and their related characteristics \cite{2009i2b2, kumari2b2}. The Evidence-Based Medicine (EBM)-NLP (EBM-NLP) \cite{nye-etal-2018-corpus} corpus has  over 5,000 abstracts of medical articles describing clinical trials annotated for \textit{population}, \textit{intervention} \textit{comparators}, and \textit{outcome} entities. This dataset is widely used for benchmarking and evaluating model performance in the biomedical domain \citep{pubmedbert2022}. The EBM-NLP corpus utilized medical experts and Amazon Mechanical Turk (MTurk) workers. The BioCreative-V Chemical Disease Relation (BC5CDR) corpus \citep{bc5cdr_Li2016} contains PubMed articles annotated for disease and chemical entities. Other corpora include annotations for biomedical phenomena like drug-drug interaction \cite{HERREROZAZO} and gene-disease associations \cite{GAD_becker2004}.  

Prior work describes the benefits of extracting information from online forums for providing better treatments to patients \cite{leaman2010towards, Gupta2014,sampathkumar2014mining, huynh2016adverse}. This information can be extracted from medical condition-specific forums (e.g. breastcancer.org, diabetesinfo etc.) or social media platforms (e.g. Reddit, Twitter). For e.g.\cite{Athira2021,romberg-etal-2020} extract medical condition-specific information from patient forums. Other works use social media platforms like Reddit to identify suicide risk \cite{zirikly-etal-2019-clpsych}, detect anxiety\cite{shen-rudzicz-2017-detecting}, and depression \cite{pirina-coltekin-2018-identifying}, and identify Schizophrenia by using linguistic analysis \cite{zomick-etal-2019-linguistic} of the posts from online discussion forums. Social media and online forum corpora have been annotated with a range of medical information.  The Self-Reported Mental Health Diagnoses (SMHD) dataset consists of Reddit posts of users who claimed to have been diagnosed with mental health conditions \cite{cohan2018smhd} and build machine-learning-based systems to automatically classify diagnosed users from control with Reddit posts. The Early risk prediction on the Internet (eRisk) \citep{parapar_eRisk} corpus focuses on detecting various text-related indicators of mental health conditions in Reddit posts. 

Medical IE tasks have been pursued using rule-based and data-driven approaches. Early medical IE work used rule-based systems with semantic and lexical information \cite{yanghuirule2010} or discrete machine learning approaches like Logistic Regression, Support Vector Machine(SVM) \cite{patrick-li-2009-cascade} with engineered features. More recent medical IE work leverages neural networks. Contemporary state-of-the-art medical IE work utilizes pre-trained language models, like BERT \cite{devlin2019bert}, which have gained prominence across various NLP tasks and biomedical applications. There are many domain-specific BERT models, including  clinical and biomedical variants with mixed domain and in-domain pre-training techniques \cite{alsentzer-etal-2019-publicly,biobert2019,bioredditbert}.

Initial work using RedHoT focused on building a machine-learning-based system to verify user claims. The RedHot \cite{redhot2022} corpus contains Reddit posts annotated for patient experiences, claims, claim-based personal experiences, questions, and granular annotations describing patient populations, interventions, and outcomes across 24 health conditions. The system extracts relevant user claims and PIO entities from user posts. The system then uses this extracted information to inform the retrieval of trustworthy evidence relevant to the claim from a scientific knowledge base.

\section{Methods}

\subsection{Data}
The corpus for the SemEval causal claims shared task was a subset of the RedHOT corpus annotated for medical condition entities across nine medical conditions.  The posts are annotated in two stages: (1) Subtask-1 annotations include user \textit{claims}, \textit{personal experiences}, \textit{questions}, and \textit{claim-based personal experiences} and (2) Subtask-2 annotations contain more granular information where entities that indicate patient \textit{population}, \textit{intervention}, and \textit{outcome} are annotated. We used the extraction script provided by the organizers to scrape the text corresponding to the postIDs off Reddit. Some user posts may be deleted on Reddit over time.  Our training set for Subtasks 1 and 2 contained 4964 posts and 517 posts, respectively. The RedHot corpus was annotated by a combination of experts and MTurk workers.  The inter-annotator agreement varies by entity type. Entities like \textit{question}, \textit{population} have higher annotator agreement (>70\% F1), and \textit{claim}, \textit{experience}, \textit{intervention}, and \textit{outcome} entities have lower annotator agreement (<55\% F1).  Table \ref{post_dist} contains the distribution of the user posts across different medical condition subreddits.

\begin{table}[ht!] 
    \small
    \centering

\begin{tabular}{lccc}\toprule
\textbf{Medical Condition} &\textbf{Subtask-1} &\textbf{Subtask-2} \\\midrule
Cystic Fibrosis &641 &55 \\
Epilepsy &415 &34 \\
GERD &384 &43 \\
Gout &630 &108 \\
IBS &575 &46 \\
Lupus &501 &54 \\
Multiple Sclerosis &641 &96 \\
POTS &590 &49 \\
Psychosis &587 &32 \\
\textbf{Total} &\textbf{4964} &\textbf{517} \\
\bottomrule
\end{tabular}

    \caption{Number of posts in the RedHot training set by medical condition.}
    \label{post_dist}
\end{table}

\begin{figure*}[ht!] 
    \centering
    \frame{\includegraphics[width=6.00in]{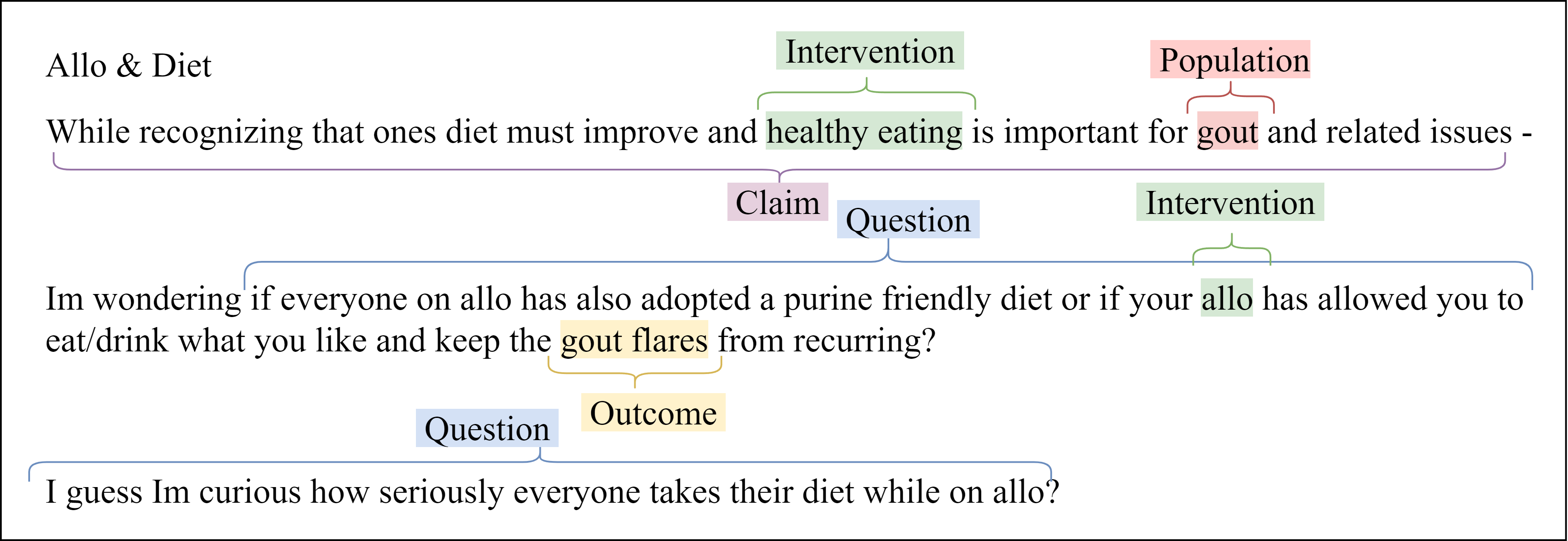}}
    \caption{Sample post from RedHot corpus annotated for Subtask-1 and Subtask-2 entities.}
    \label{annotation_sample}
\end{figure*}

Users on medical condition-related subreddits often discuss their challenges dealing with the symptoms and medications and seek advice from other users who may have had similar experiences. Users may also provide advice based on their experience dealing with recovery or progress after going through treatment. These claims often describe a treatment/intervention (medical and non-medical) that affects patient outcomes related to the medical condition. \textit{Personal experience} could describe the user's experience, e.g., the trajectory of their condition when exposed to interventions. User posts that describe claims in a personal context are annotated as \textit{claim personal experience}. Figure \ref{annotation_sample} contains a post from the RedHot corpus annotated for the relevant entities.

\begin{table}[ht!] 
    \small
    \centering

\begin{tabular}{p{3mm}p{35mm}p{12mm}p{13mm}}\toprule
&\textbf{Entity} &\textbf{Frequency} &\textbf{\# Tokens} \\\midrule
\multirow{4}{*}{1} &Claim &462 &19.6 \\
&Claim-Personal-Experience &{1198} &30.6 \\
&Personal Experience&{5744} &27.0 \\
&Question &5363 &10.5 \\
\midrule
\multirow{3}{*}{2} &Population &396 &1.2 \\
&Intervention &526 &1.6 \\
&Outcome &452 &1.9 \\
\bottomrule
\end{tabular}

    \caption{Frequency distribution of entities in the training set and their average lengths (in tokens) for Subtasks 1 and 2 }
    \label{entity_distribution}
\end{table}

Table \ref{entity_distribution} contains the entity distribution and average length of the entities from Subtask-1 and -2 annotations for the RedHot training set. The \textit{personal experience} and \textit{question} entities are more frequent than \textit{claims} and \textit{claim-based personal experiences} entities. \textit{questions} tend to be short, and text describing personal experiences is relatively longer. \textit{claim-based personal experiences} contain additional text that describes the user's knowledge about an \textit{intervention}, making them longer than \textit{claims} or \textit{personal experiences}. 

Entities from the Subtask-2 annotation tend to be shorter. \textit{population} and \textit{outcome} may describe a medical condition or an effect of having the condition. \textit{Intervention} spans could be medical and non-medical text that indicates a treatment, procedure, or other actions users may have taken while undergoing a described experience. A user post rarely contains more than one \textit{claim} (<1\%). Across all entity types, overlapping entity spans are rare (<1\%) in the training set.

\subsection{Evaluation}

The organizers framed both subtasks as sequence tagging tasks, and a prediction was required for every test set token (whitespace word tokenization was required by the organizers). For test set evaluation, Subtask-1 was evaluated at the sentence level, and Subtask-2 was evaluated for token-level Precision (P), Recall (R), and F1 scores. Due to data-use restrictions, participants could not access the test set labels or the evaluation scripts. Apart from the performance reported on the task leaderboards, we additionally report performance on a held-out validation set and perform detailed error analysis on the validation set due to the above-mentioned data and evaluation restrictions. For the validation set evaluation, we evaluate our models using token-level P, R, and F1 scores following \cite{redhot2022}. We validate the effectiveness of the systems using a strict non-parametric (bootstrap) test \cite{bergkirkpatrick2012}.

\subsection{Medical-condition information extraction}

Subtask-1 annotation tend to be full sentences and can be conceptualized as either a sentence classification task or a sequence tagging task where each sentence/span can be labeled as - \textit{claim}, \textit{personal experience}, \textit{claim personal experience}, and \textit{question}. We experimented with both of these conceptualizations but found no significant difference in performance between the two approaches. We approach Subtask-2 as an entity extraction task. We fine-tuned general and mixed-domain language models and studied the inclusion of external knowledge about the entities of interest into these language models.

\textbf{Fine-tuning General and Mixed-domain pre-trained models:} Language models are pre-trained on the source domain and then fine-tuned to a specific task in a target domain. Prior work \cite{dont_stop2020,bioredditbert} indicates that language models benefit from performing unsupervised learning tasks on target domains. Additional pre-training on the text more similar to the target domain typically improves performance when the original source and target domains are more dissimilar \cite{dont_stop2020,domain_generalization2022}. To study this, we fine-tuned three BERT models that have been pre-trained on general and mixed-domain corpora. We selected \textit{BERT} \citep{devlin2019bert} and \textit{RoBERTa} \cite{Roberta2019} as our general-domain variants. BERT is trained on a corpus of Wikipedia articles and books, while RoBERTa is trained on a larger web crawl corpus. Our mixed-domain variants include \textit{BioMedRoBERTa} \cite{dont_stop2020} 
 and \textit{BioRedditBERT} \cite{bioredditbert}. BioMedRoBERTa is the RoBERTa model subsequently pre-trained on the S2ORC corpus \cite{Lo2019GORCAL} and the BioRedditBERT model is the BioBERT \cite{biobert2019} subsequently pre-trained on health-related posts from Reddit.

\begin{figure*}[ht!]
    \centering
    \frame{\includegraphics[width=6.00in]{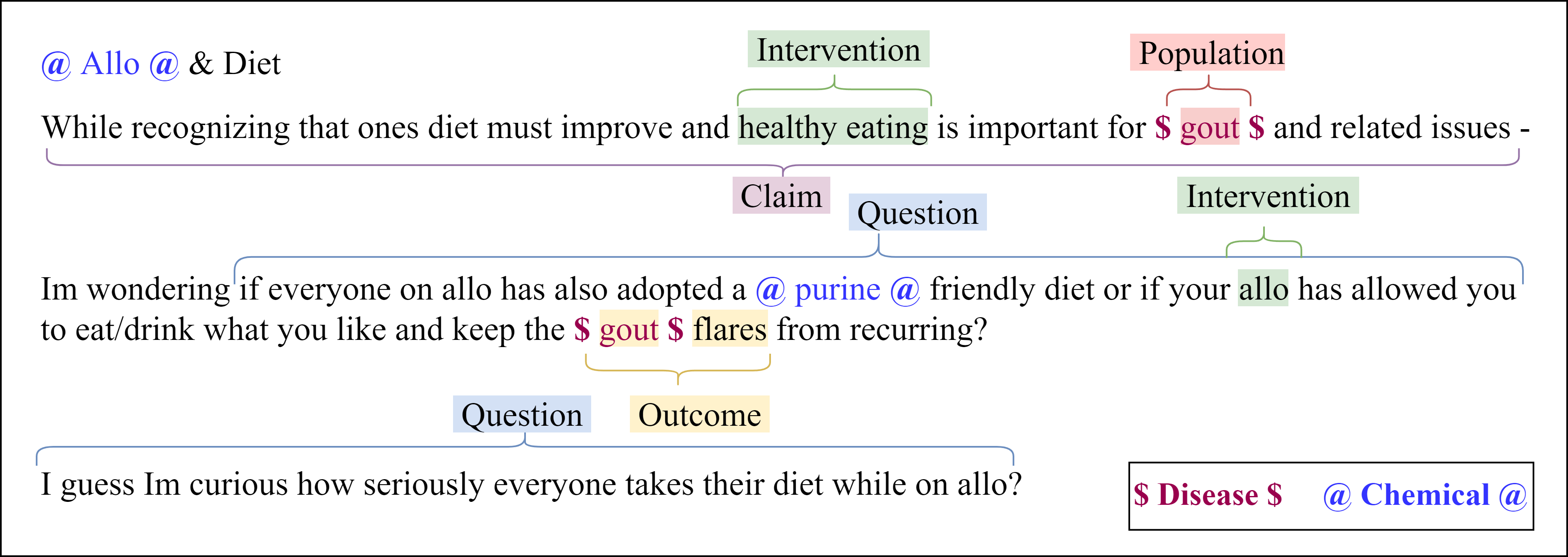}}
    \caption{A post from the RedHot training set with knowledge augmentations for Disease and Chemical entities. We use a model trained on the BioCreative V corpus to predict the Disease and Chemical entities and encode them with special tokens `\$\$' and `@@'.}
    \label{data_augmentation}
\end{figure*}

\textbf{Utilizing relevant external knowledge:} 
Prior work \cite{roy-pan-2021-incorporating,harnoune2021} found performance improvements for target domain tasks by incorporating relevant knowledge in BERT-based methods via knowledge graphs. We incorporate external knowledge through pre-trained models via data augmentation. Our data analysis observed some overlap between disease names and the \textit{population} entities and \textit{intervention} entities indicating usage of medications. To analyze the overlap, we randomly sampled ~50 posts that contained 35 \textit{population} entities and 68 \textit{interventions}. We found that 31 out of 35 \textit{population} entities were disease names or their abbreviations, such as ``gout,'' ``MS,'' ``IBS,'' ``CF,'' ``inflammation,'' etc., and 43 out of 68 spans labeled \textit{interventions} were chemical names such as ``allopurinol,'' ``metoclopramide,'' ``hydroxychloroquine'', etc.. We used the \textit{en\_ner\_bc5cdr\_md}, a  pre-trained model \cite{scispacy2019} trained on BC5CDR corpus \cite{bc5cdr_Li2016} for incorporating external knowledge. The disease-chemical model is trained on over 1500 PubMed articles with disease and chemical entity annotations. Inspired by prior work on data augmentation using special tokens \cite{zhong2021frustratingly,RAMACHANDRAN2023104302}, we augment the given user posts in training and inference to encode disease and chemical entities with special tokens (\$\$ and `@@') based on predictions from the pre-trained BC5CDR model. A detailed illustration of the data augmentation is presented in Figure \ref{data_augmentation}. 

We train separate models with identical architectures for extracting the Subtask-1 and Subtask-2 entities using general and mixed-domain BERT models and perform knowledge enhancement via data augmentation on these BERT models. We implement BERT-based sequence tagging models by adding a linear output layer to the BERT hidden state using a begin-inside-outside (BIO) label. 
 
\subsection{Experimental setup}

We split the provided training set for each Subtask-1 and -2 into new training sets and validation sets for model development and parameter tuning. We sampled the Subtask-1 and -2 validation sets intentionally biased to reflect the entity distribution in the provided data (as in Table \ref{entity_distribution}). All our models operated at a sentence level, and we tokenized the sentences in the user posts using scispaCy \cite{scispacy2019} \textit{en\_core\_sci\_md} tokenizer. The systems were implemented in PyTorch using the HuggingFace transformers library \cite{wolf-etal-2020-transformers}. We used AdamW optimizer \cite{adamw2017}. We used grid search and tuned the hyperparameters- epochs and batch size on the validation set. We left the dropout and learning rate at recommended default values. The final hyperparameters include a training batch size of 64, a validation batch size of 16, a maximum sequence length of 256, a dropout rate of 0.2 for regularizing the model, and the initial learning rate at 5e-5. We trained the models for Subtask-1 for ten epochs, while the models for Subtask-2 were trained for 20 epochs.

\section{Results and Discussion}

\subsection{Test set performance}
Table \ref{testset_1} and \ref{test_set_2} present the test set performances for both subtasks, respectively. We applied BERT-base for both Subtask-1 and -2. We present our system's performance (ranked 3\textsuperscript{rd} in both subtasks) along with the performance of top-2 teams and the median performance for comparison.
For Subtask-1, the median score is the median of the seven system performances posted on the task leaderboard, and for Subtask-2, the median score presented is the median of the six system performances posted on the task leaderboard.

\begin{table}[ht!]  
    \small
    \centering

\begin{tabular}{lrrrr}\toprule
&\textbf{P} &\textbf{R} &\textbf{F1} \\\midrule
Team 1 &78.14 &78.65 &78.40 \\
Team 2 &72.97 &67.36 &70.05 \\
\textbf{Ours} &71.16 &65.78 &68.59 \\
Median &70.24 &62.90 &65.70 \\
\bottomrule
\end{tabular}
    \caption{Subtask-1: Test set performance of the top-3 teams and the median performance. The systems were evaluated at a sentence level using Precision(P), Recall (R), and F1 scores. The median score is the median of the seven system performances posted on the task leaderboard.  }
    \label{testset_1}
\end{table}

\begin{table}[ht!]  
    \small
    \centering

\begin{tabular}{lrrrr}\toprule
&\textbf{Population} &\textbf{Intervention} &\textbf{Outcome} \\\midrule
Team 1 &40.55 &49.71 &30.08 \\
Team 2 &37.78 &43.58 &30.67 \\
\textbf{Ours} &34.96 &42.16 &20.83 \\
Median &33.59 &41.72 &23.27 \\
\bottomrule
\end{tabular}

    \caption{Subtask-2: Test set performance of the top-3 teams and the median performance. The systems were evaluated using token-level F1 scores. The median score is the median of the six system performances posted on the task leaderboard. }
    \label{test_set_2}
\end{table}

 \begin{table*}[ht!]  
    \small
    \centering

\begin{tabular}{lccccccc}\toprule
\multirow{2}{*}{\textbf{Model}} &\multirow{2}{*}{\textbf{Metric}} &\multirow{2}{*}{\textbf{Claim}} &\textbf{Claim Personal} &\textbf{Personal} &\multirow{2}{*}{\textbf{Question}} &\multirow{2}{*}{\textbf{Overall}} \\
& & &\textbf{Experience} &\textbf{Experience} & & \\
\textbf{} &\textbf{} &n=2307 &n=8315 & n=35280 & n=12693 & \\\midrule
\multirow{3}{*}{BERT-base\textsuperscript{$\delta$}} &\textbf{P} &30.93 &28.52 &51.54 &80.81 &47.95 \\
&\textbf{R} &28.05 &26.05 &59.45 &81.97 &48.88 \\
&\textbf{F1} &\textbf{29.42} &27.23 &55.21 &81.39 &48.31 \\
\midrule
\multirow{3}{*}{BioMedRoBERTa} &\textbf{P} &39.63 &33.40 &52.22 &80.22 &51.37 \\
&\textbf{R} &19.55 &26.53 &\textbf{62.09} &\textbf{86.16} &48.58 \\
&\textbf{F1} &26.18 &29.57 &\textbf{56.73} &83.08 &48.89 \\
\midrule
\multirow{3}{*}{BioRedditBERT} &\textbf{P} &21.14 &28.07 &\textbf{54.96} &\textbf{82.01} &46.55 \\
&\textbf{R} &\textbf{35.41} &\textbf{37.08} &51.23 &84.66 &\textbf{52.10} \\
&\textbf{F1} &26.48 &\textbf{31.95} &53.03 &\textbf{83.32} &48.69 \\
\midrule
\midrule
\multirow{3}{*}{BERT-base+} &\textbf{P} &31.18 &31.69 &50.47 &78.85 &48.05 \\
&\textbf{R} &24.27 &22.87 &60.08 &85.39 &48.15 \\
&\textbf{F1} &27.30 &26.56 &54.86 &81.99 &47.68 \\\midrule
\multirow{3}{*}{BioMedRoBERTa+} &\textbf{P} &29.31 &31.86 &53.34 &81.36 &48.97 \\
&\textbf{R} &22.71 &26.98 &55.92 &81.18 &46.70 \\
&\textbf{F1} &25.59 &29.22 &54.60 &81.27 &47.67 \\
\midrule
\multirow{3}{*}{BioRedditBERT+} &\textbf{P} &\textbf{41.53} &\textbf{42.15} &54.43 &79.59 &\textbf{54.43} \\
&\textbf{R} &22.76 &23.62 &58.62 &84.42 &47.35 \\
&\textbf{F1} &29.40 &30.28 &56.45 &81.93 &\textbf{49.52} \\
\bottomrule
\end{tabular}
    \caption{Subtask-1: P, R and F1 scores at the token level on the validation set for patient experience-related entities in RedHot. Models with '+' were augmented with external knowledge of Disease and Chemical knowledge. $\delta$- we used the BERT-base system to predict the entity types in the test set.}
    \label{experiments_1}
\end{table*}

 \begin{table*}[ht!]  
    \small
    \centering

\begin{tabular}{lcccccc}\toprule
\multirow{2}{*}{\textbf{Model}} &\multirow{2}{*}{\textbf{Metric}} &\textbf{Population} &\textbf{Intervention} &\textbf{Outcome} &\multirow{2}{*}{\textbf{Overall}} \\
& &n=106 &n=206 &n=168 & \\\midrule
\multirow{3}{*}{BERT-base\textsuperscript{$\delta$}} &\textbf{P} &18.57 &23.27 &14.14 &18.66 \\
&\textbf{R} &24.53 &27.67 &16.57 &22.92 \\
&\textbf{F1} &21.14 &25.28 &15.26 &20.56 \\\midrule
\multirow{3}{*}{BioMedRoBERTa} &\textbf{P} &23.38 &18.51 &13.72 &18.53 \\
&\textbf{R} &33.96 &30.10 &18.34 &27.47 \\
&\textbf{F1} &27.69 &22.92 &15.70 &22.10 \\\midrule
\multirow{3}{*}{BioRedditBERT} &\textbf{P} &15.70 &24.72 &14.41 &18.28 \\
&\textbf{R} &35.85 &32.04 &20.12 &29.34 \\
&\textbf{F1} &21.84 &27.91 &16.79 &22.18 \\\midrule \midrule
\multirow{3}{*}{BERT-base+} &\textbf{P} &\textbf{32.56} &21.76 &\textbf{19.05} &\textbf{24.46} \\
&\textbf{R} &\textbf{39.62} &22.60 &16.57 &26.26 \\
&\textbf{F1} &\textbf{35.74} &22.17 &\textbf{17.72} &25.21 \\\midrule
\multirow{3}{*}{BioMedRoBERTa+} &\textbf{P} &28.85 &21.75 &11.49 &20.70 \\
&\textbf{R} &28.30 &\textbf{37.02} &15.98 &27.10 \\
&\textbf{F1} &28.57 &27.40 &13.37 &23.11 \\\midrule
\multirow{3}{*}{BioRedditBERT+} &\textbf{P} &27.27 &\textbf{34.18} &11.90 &24.45 \\
&\textbf{R} &\textbf{39.62} &32.21 &\textbf{26.63} &\textbf{32.82} \\
&\textbf{F1} &32.31 &\textbf{33.17} &16.45 &\textbf{27.31}\textsuperscript{*} \\
\bottomrule
\end{tabular}
    \caption{Subtask-2:P, R and F1 scores at the token-level on the validation set for medical condition  entities in RedHot. Models with '+' were augmented with external knowledge of Disease and Chemical names. '*' indicates performance significance ($p<0.05$) compared to the non-augmented mixed domain and general domain models- (BERT-base, BioMedRoBERTa, and BioRedditBERT) $\delta$- we used the BERT-base system to predict the entity types in the test set.}
    \label{experiments_2}
\end{table*}

\subsection{Validation set performance}
To provide a more granular performance breakdown and a more comprehensive set of the developed architectures, we additionally report the performance of our systems on held-out validation sets and perform a detailed error analysis of our model predictions. Table  \ref{experiments_1} and \ref{experiments_2} contains the performance for extracting patient experience-related and medical condition-related entities on the validation sets, evaluated at the token level. We present the performance of six models in total. BERT-base and RoBERTa models showed similar performance; hence we only tabulate BERT-base's performance. Models with a '+' next to their names were trained on knowledge-augmented training data as described in Figure \ref{data_augmentation}.
\begin{table*}[ht!]
\small
\centering
\begin{tabular}{lrcccccc}\toprule
&\multicolumn{6}{c}{\textbf{Predicted}} \\
\multirow{6}{*}{\textbf{Gold}} & &Claim &Claim Pers.Exp &Pers.Exp &Question &No label \\\midrule
&Claim &525 &30 &170 &75 &1507 \\
&Claim Pers.Exp &37 &1964 &2762 &86 &3466 \\
&Pers.Exp &26 &968 &20681 &403 &13202 \\
&Question &37 &19 &373 &10715 &1549 \\
&No label &639 &1678 &14009 &2184 &61036 \\
\bottomrule
\end{tabular}
\caption{Token-level confusion matrix for Subtask-1 using the predictions from our best performing model, \textit{BioRedditBERT+} on the validation set.}
\label{cm1}
\end{table*}

\begin{table*}[ht!]
\small
\centering
\begin{tabular}{lrcccc}\toprule
&\multicolumn{5}{c}{\textbf{Predicted}} \\
\multirow{5}{*}{\textbf{Gold}} & &Population &Intervention &Outcome &No label \\\midrule
&Population &42 &1 &12 &51 \\
&Intervention &4 &67 &1 &136 \\
&Outcome &4 &0 &45 &120 \\
&No label &104 &128 &320 &18269 \\
\bottomrule

\end{tabular}
\caption{Token-level confusion matrix for Subtask-2 using the predictions from our best performing model, \textit{BioRedditBERT+} validation set.}
\label{cm2}
\end{table*}
Examining the results for Subtask-1, we can see that general domain language models (such as BERT) and domain-specific models (such as BioRedditBERT and BioMedRoBERTa) exhibit similar overall performance. This suggests that incorporating external knowledge about diseases and chemicals did not significantly improve for longer entity spans, such as the user's personal experiences and disease-related questions. This may be explained due to (1) the low annotator agreements, (2) ambiguity in annotated phenomena, like \textit{claim}, \textit{claim personal experience}  (3) augmented external knowledge may not be sufficient to disambiguate the entities in the Subtask-1 annotation. We provide detailed examples of prediction errors in the following section. 
 
The \textit{BioRedditBERT+} model performs significantly better ($p<0.05$) overall across all the other models and specifically when extracting \textit{population}, showing the benefits of using both a domain-specific language model and augmenting disease knowledge. The overall performance among the knowledge-augmented models did not significantly differ. The annotated phenomena in Subtask-2 (PIO) are shorter spans that describe disease or medication names that have direct overlap with the external knowledge we augmented with. We used the BERT-base system (indicated by $\delta$ in Table \ref{experiments_1} ) to predict the entity types in the test set for Subtask-1 since it did not significantly differ in performance from the other systems. For Subtask-2, we decided to again predict the entity types in the test set using BERT-base (indicated by $\delta$ in Table \ref{experiments_2}). We believe that the proposed domain-specific knowledge-augmented systems will perform significantly better in practical use when extracting these medical condition-related entities.

\subsection{Error Analysis:}
We performed a detailed error analysis on the validation sets for Subtask-1 and -2.  Given the relatively low F1 scores across PIO entity types, we focused the error analysis on the mislabeled samples by all models.  We include Tables \ref{cm1} and \ref{cm2} containing the token-level confusion matrices to understand the trends in misclassifications by our system across Subtask-1 and Subtask-2. We discuss these trends with specific examples from the posts in the validation set in the sections below for better understanding.

\subsubsection{Errors in extracting patient experience-related entities}

\textbf{User Question descriptions:} \textit{Question} has higher performance compared to all the other entities, which is likely attributable to \textit{question} spans tend to be shorter, typically start with a question word, and frequently end with a question mark. However, we observed users often use question-like sentences to describe \textit{claim} or \textit{claim personal experience}, which were harder for models to predict. %example

\textbf{Sentence tokenization:} Some errors were due to the sentence tokenization. While models could correctly predict longer entities, they often failed to predict longer sequences spanning multiple sentences. For example, the description  ``Pantoprazole intake for the past two weeks since have been awfully off, to say the least: ..., but I feel like my chest has to work for it more'',  with >75 tokens, annotated as \textit{claim personal experience} was correctly predicted whereas the description, ``The medical doctor says I dont have it. Naturopathic doctors says I have it. ...Medical doc says that is just IBS'' annotated as \textit{claim personal experience} with ~50 tokens was split into five sentences and misclassified. These errors may be resolved by including cross-sentence context.

\textbf{Confusability between experience-related entities:} Entity types \textit{claim}, \textit{claim personal experience} and \textit{personal experience} were confusable. For example, the description, ``I'm T1D and my blood sugars went low.,'' was annotated as \textit{claim} and predicted by all the systems as \textit{personal experience}.  Similarly, distinguishing \textit{claim personal experience} and \textit{personal experience} is challenging due to the closeness in the structure. For example, ``Now the Gouts gone my Ankle is grade 3 sprained from walking funny and putting too much pressure on it'' while the gold label is \textit{claim personal experience}, it was predicted as \textit{personal experience}. In another instance, ``I took them but my pain and swelling got worst during night.,'' was annotated as \textit{personal experience} and predicted as \textit{claim personal experience}.

%Stage-2

\subsubsection{Errors in PIO extraction}
Five frequent medical conditions - ``Gout,'' ``Multiple Sclerosis,'' ``POTS,'' ``Lupus,'' and ``Cystic Fibrosis,'' account for  ~35\% of the  \textit{population} entity phrases in the validation set. These phrases were extracted with high performance by domain-specific knowledge-augmented systems compared to the general-domain models without external knowledge. Abbreviated medical conditions were challenging classification targets or were confused between \textit{population} and \textit{outcome} when abbreviations appear in different contexts. Users sometimes coreference people suffering from the medical condition using phrases like ``patients'' or ``individuals.'' These were annotated as \textit{population} but were not predicted by the models. 
Some \textit{intervention} phrases mention disease names followed by ``drugs'', e.g.  ``Cystic Fibrosis drugs'' or ``seizure medication''. The medical conditions in these phrases were predicted as \textit{population} and words following these disease names- ``drugs'' and ``medication'' were predicted as \textit{intervention}. Some false negative \textit{intervention} phrases included brand names (e.g. ``pepsi'') or informal names for medications (e.g. ``Tobi'' for the medication ``Tobramycin'').

\section{Conclusions}

We explore a novel medical information extraction task in which Reddit posts are characterized by entities related to patient experiences and medical conditions. We  perform this IE task from a sequence-tagging approach using domain-specific and knowledge-augmented systems. Domain-specific pre-trained models are better at identifying the medical condition-related entities while general-domain pre-trained models are as good as domain-specific models in understanding patient experience-related descriptions. The inclusion of external knowledge, specifically disease information, helped improve \textit{population} entity detection, and including chemical information helped identify \textit{intervention}. With thorough error analysis, we identify where knowledge-augmented systems overcome errors faced by general-domain systems. In future work, it may be useful to utilize cross-sentence information and perform entity normalization to extract the experience-related and medical condition-related entities with higher performance.

\bibliography{custom}

\begin{thebibliography}{39}
\expandafter\ifx\csname natexlab\endcsname\relax\def\natexlab#1{#1}\fi

\bibitem[{Alsentzer et~al.(2019)Alsentzer, Murphy, Boag, Weng, Jin, Naumann,
  and McDermott}]{alsentzer-etal-2019-publicly}
Emily Alsentzer, John Murphy, William Boag, Wei-Hung Weng, Di~Jin, Tristan
  Naumann, and Matthew McDermott. 2019.
\newblock \href {https://doi.org/10.18653/v1/W19-1909} {Publicly available
  clinical {BERT} embeddings}.
\newblock In \emph{Clinical Natural Language Processing Workshop}, pages
  72--78.

\bibitem[{Athira et~al.(2021)Athira, Jones, Idicula, Kulanthaivel, and
  Zhang}]{Athira2021}
B.~Athira, Josette Jones, Sumam~Mary Idicula, Anand Kulanthaivel, and Enming
  Zhang. 2021.
\newblock \href {https://doi.org/10.1186/s40537-021-00429-7} {Annotating and
  detecting topics in social media forum and modelling the annotation to derive
  directions-a case study}.
\newblock \emph{Journal of Big Data}, 8(1).

\bibitem[{Basaldella et~al.(2020)Basaldella, Liu, Shareghi, and
  Collier}]{bioredditbert}
Marco Basaldella, Fangyu Liu, Ehsan Shareghi, and Nigel Collier. 2020.
\newblock \href {https://doi.org/10.48550/ARXIV.2010.03295} {Cometa: A corpus
  for medical entity linking in the social media}.

\bibitem[{Becker et~al.(2004)Becker, Barnes, Bright, and Wang}]{GAD_becker2004}
Kevin~G Becker, Kathleen~C Barnes, Tiffani~J Bright, and S~Alex Wang. 2004.
\newblock \href {https://www.nature.com/articles/ng0504-431} {The genetic
  association database}.
\newblock \emph{Nat. Genet.}, 36(5):431--432.

\bibitem[{Berg-Kirkpatrick et~al.(2012)Berg-Kirkpatrick, Burkett, and
  Klein}]{bergkirkpatrick2012}
Taylor Berg-Kirkpatrick, David Burkett, and Dan Klein. 2012.
\newblock \href {https://aclanthology.org/D12-1091} {An empirical investigation
  of statistical significance in {NLP}}.
\newblock In \emph{Proceedings of the 2012 Joint Conference on Empirical
  Methods in Natural Language Processing and Computational Natural Language
  Learning}, pages 995--1005, Jeju Island, Korea. Association for Computational
  Linguistics.

\bibitem[{Cohan et~al.(2018)Cohan, Desmet, Yates, Soldaini, MacAvaney, and
  Goharian}]{cohan2018smhd}
Arman Cohan, Bart Desmet, Andrew Yates, Luca Soldaini, Sean MacAvaney, and
  Nazli Goharian. 2018.
\newblock \href {https://www.aclweb.org/anthology/C18-1126} {Smhd: A
  large-scale resource for exploring online language usage for multiple mental
  health conditions}.
\newblock In \emph{Proceedings of the 27th International Conference on
  Computational Linguistics (COLING)}, pages 1485–--1497. Association for
  Computational Linguistics.

\bibitem[{Devlin et~al.(2019)Devlin, Chang, Lee, and
  Toutanova}]{devlin2019bert}
Jacob Devlin, Ming Chang, Kenton Lee, and Kristina Toutanova. 2019.
\newblock \href {https://doi.org/10.18653/v1/N19-1423} {{BERT}: Pre-training of
  deep bidirectional transformers for language understanding}.
\newblock In \emph{North American Chapter of the Association for Computational
  Linguistics}, pages 4171--4186.

\bibitem[{Gu et~al.(2022)Gu, Tinn, Cheng, Lucas, Usuyama, Liu, Naumann, Gao,
  and Poon}]{pubmedbert2022}
Yu~Gu, Robert Tinn, Hao Cheng, Michael Lucas, Naoto Usuyama, Xiaodong Liu,
  Tristan Naumann, Jianfeng Gao, and Hoifung Poon. 2022.
\newblock \href {https://doi.org/10.1145/3458754} {Domain-specific language
  model pretraining for biomedical natural language processing}.
\newblock \emph{ACM Transactions on Computing for Healthcare}, 3(1):1–23.

\bibitem[{Gupta et~al.(2014)Gupta, MacLean, Heer, and Manning}]{Gupta2014}
Sonal Gupta, Diana~L MacLean, Jeffrey Heer, and Christopher~D Manning. 2014.
\newblock \href {https://doi.org/10.1136/amiajnl-2014-002669} {{Induced
  lexico-syntactic patterns improve information extraction from online medical
  forums}}.
\newblock \emph{Journal of the American Medical Informatics Association},
  21(5):902--909.

\bibitem[{Gururangan et~al.(2020)Gururangan, Marasovic, Swayamdipta, Lo,
  Beltagy, Downey, and Smith}]{dont_stop2020}
Suchin Gururangan, Ana Marasovic, Swabha Swayamdipta, Kyle Lo, Iz~Beltagy, Doug
  Downey, and Noah~A. Smith. 2020.
\newblock \href {http://arxiv.org/abs/2004.10964} {Don't stop pretraining:
  Adapt language models to domains and tasks}.
\newblock \emph{CoRR}, abs/2004.10964.

\bibitem[{Harnoune et~al.(2021)Harnoune, Rhanoui, Mikram, Yousfi, Elkaimbillah,
  and {El Asri}}]{harnoune2021}
Ayoub Harnoune, Maryem Rhanoui, Mounia Mikram, Siham Yousfi, Zineb
  Elkaimbillah, and Bouchra {El Asri}. 2021.
\newblock \href {https://doi.org/https://doi.org/10.1016/j.cmpbup.2021.100042}
  {{BERT} based clinical knowledge extraction for biomedical knowledge graph
  construction and analysis}.
\newblock \emph{Computer Methods and Programs in Biomedicine Update}, 1:100042.

\bibitem[{Herrero-Zazo et~al.(2013)Herrero-Zazo, Segura-Bedmar, Martínez, and
  Declerck}]{HERREROZAZO}
María Herrero-Zazo, Isabel Segura-Bedmar, Paloma Martínez, and Thierry
  Declerck. 2013.
\newblock \href {https://doi.org/https://doi.org/10.1016/j.jbi.2013.07.011}
  {The ddi corpus: An annotated corpus with pharmacological substances and
  drug–drug interactions}.
\newblock \emph{Journal of Biomedical Informatics}, 46(5):914--920.

\bibitem[{Hui(2010)}]{yanghuirule2010}
Yang Hui. 2010.
\newblock \href {https://doi.org/10.1136/jamia.2010.003863} {Automatic
  extraction of medication information from medical discharge summaries}.
\newblock \emph{J Am Med Inform Assoc JAMIA}, 17,5:545–548.

\bibitem[{Huynh et~al.(2016)Huynh, He, Willis, and Rueger}]{huynh2016adverse}
Trung Huynh, Yulan He, Alistair Willis, and Stefan Rueger. 2016.
\newblock \href {https://aclanthology.org/C16-1084} {Adverse drug reaction
  classification with deep neural networks}.
\newblock In \emph{Proceedings of {COLING} 2016, the 26th International
  Conference on Computational Linguistics: Technical Papers}, pages 877--887,
  Osaka, Japan. The COLING 2016 Organizing Committee.

\bibitem[{Khetan et~al.(2023)Khetan, Wadhwa, Wallace, and
  Amir}]{khetan-EtAl:2023:SemEval}
Vivek Khetan, Somin Wadhwa, Byron Wallace, and Silvio Amir. 2023.
\newblock Semeval-2023 task 8: Causal medical claim identification and related
  pio frame extraction from social media posts.
\newblock In \emph{Proceedings of the 17th International Workshop on Semantic
  Evaluation}, Toronto, Canada. Association for Computational Linguistics.

\bibitem[{Kumar et~al.(2015)Kumar, Stubbs, Shaw, and Uzuner}]{kumari2b2}
Vishesh Kumar, Amber Stubbs, Stanley Shaw, and {\"O}zlem Uzuner. 2015.
\newblock \href {https://doi.org/10.1016/j.jbi.2015.09.018} {Creation of a new
  longitudinal corpus of clinical narratives}.
\newblock \emph{J Biomed Inform}, 58S.

\bibitem[{Leaman et~al.(2010)Leaman, Wojtulewicz, Sullivan, Skariah, Yang, and
  Gonzalez}]{leaman2010towards}
Robert Leaman, Laura Wojtulewicz, Ryan Sullivan, Annie Skariah, Jian Yang, and
  Graciela Gonzalez. 2010.
\newblock \href {https://dl.acm.org/doi/10.5555/1869961.1869976} {Towards
  internet-age pharmacovigilance: extracting adverse drug reactions from user
  posts to health-related social networks}.
\newblock In \emph{Proceedings of the 2010 workshop on biomedical natural
  language processing}, pages 117--125.

\bibitem[{Lee et~al.(2019)Lee, Yoon, Kim, Kim, Kim, So, and Kang}]{biobert2019}
Jinhyuk Lee, Wonjin Yoon, Sungdong Kim, Donghyeon Kim, Sunkyu Kim, Chan~Ho So,
  and Jaewoo Kang. 2019.
\newblock \href {https://doi.org/10.1093/bioinformatics/btz682} {{BioBERT}: a
  pre-trained biomedical language representation model for biomedical text
  mining}.
\newblock \emph{Bioinformatics}.

\bibitem[{Li et~al.(2016)Li, Sun, Johnson, Sciaky, Wei, Leaman, Davis,
  Mattingly, Wiegers, and Lu}]{bc5cdr_Li2016}
Jiao Li, Yueping Sun, Robin~J Johnson, Daniela Sciaky, Chih-Hsuan Wei, Robert
  Leaman, Allan~Peter Davis, Carolyn~J Mattingly, Thomas~C Wiegers, and Zhiyong
  Lu. 2016.
\newblock \href {https://www.ncbi.nlm.nih.gov/pmc/articles/PMC4860626/}
  {{BioCreative} {V} {CDR} task corpus: a resource for chemical disease
  relation extraction}.
\newblock \emph{Database (Oxford)}, 2016:baw068.

\bibitem[{Liu et~al.(2019)Liu, Ott, Goyal, Du, Joshi, Chen, Levy, Lewis,
  Zettlemoyer, and Stoyanov}]{Roberta2019}
Yinhan Liu, Myle Ott, Naman Goyal, Jingfei Du, Mandar Joshi, Danqi Chen, Omer
  Levy, Mike Lewis, Luke Zettlemoyer, and Veselin Stoyanov. 2019.
\newblock \href {https://doi.org/10.48550/ARXIV.1907.11692} {Roberta: A
  robustly optimized {BERT} pretraining approach}.

\bibitem[{Lo et~al.(2019)Lo, Wang, Neumann, Kinney, and Weld}]{Lo2019GORCAL}
Kyle Lo, Lucy~Lu Wang, Mark Neumann, Rodney~Michael Kinney, and Daniel~S. Weld.
  2019.
\newblock Gorc: A large contextual citation graph of academic papers.
\newblock \emph{ArXiv}, abs/1911.02782.

\bibitem[{Loshchilov and Hutter(2017)}]{adamw2017}
Ilya Loshchilov and Frank Hutter. 2017.
\newblock \href {http://arxiv.org/abs/1711.05101} {Fixing weight decay
  regularization in adam}.
\newblock \emph{CoRR}, abs/1711.05101.

\bibitem[{Neumann et~al.(2019)Neumann, King, Beltagy, and Ammar}]{scispacy2019}
Mark Neumann, Daniel King, Iz~Beltagy, and Waleed Ammar. 2019.
\newblock \href {https://doi.org/10.18653/v1/w19-5034} {Scispacy: Fast and
  robust models for biomedical natural language processing}.
\newblock \emph{Proceedings of the 18th BioNLP Workshop and Shared Task}.

\bibitem[{Nye et~al.(2018)Nye, Li, Patel, Yang, Marshall, Nenkova, and
  Wallace}]{nye-etal-2018-corpus}
Benjamin Nye, Junyi~Jessy Li, Roma Patel, Yinfei Yang, Iain Marshall, Ani
  Nenkova, and Byron Wallace. 2018.
\newblock \href {https://doi.org/10.18653/v1/P18-1019} {A corpus with
  multi-level annotations of patients, interventions and outcomes to support
  language processing for medical literature}.
\newblock In \emph{Proceedings of the 56th Annual Meeting of the Association
  for Computational Linguistics (Volume 1: Long Papers)}, pages 197--207,
  Melbourne, Australia. Association for Computational Linguistics.

\bibitem[{Parapar et~al.(2022)Parapar, Mart\'{\i}n-Rodilla, Losada, and
  Crestani}]{parapar_eRisk}
Javier Parapar, Patricia Mart\'{\i}n-Rodilla, David~E. Losada, and Fabio
  Crestani. 2022.
\newblock \href {https://doi.org/10.1007/978-3-031-13643-6_18} {Overview of
  erisk 2022: Early risk prediction on the internet}.
\newblock In \emph{Experimental IR Meets Multilinguality, Multimodality, and
  Interaction: 13th International Conference of the CLEF Association, CLEF
  2022, Bologna, Italy, September 5–8, 2022, Proceedings}, page 233–256,
  Berlin, Heidelberg. Springer-Verlag.

\bibitem[{Patrick and Li(2009)}]{patrick-li-2009-cascade}
Jon Patrick and Min Li. 2009.
\newblock \href {https://aclanthology.org/U09-1014} {A cascade approach to
  extracting medication events}.
\newblock In \emph{Proceedings of the Australasian Language Technology
  Association Workshop 2009}, pages 99--103, Sydney, Australia.

\bibitem[{Pirina and
  {\c{C}}{\"o}ltekin(2018)}]{pirina-coltekin-2018-identifying}
Inna Pirina and {\c{C}}a{\u{g}}r{\i} {\c{C}}{\"o}ltekin. 2018.
\newblock \href {https://doi.org/10.18653/v1/W18-5903} {Identifying depression
  on {R}eddit: The effect of training data}.
\newblock In \emph{Proceedings of the 2018 {EMNLP} Workshop {SMM}4{H}: The 3rd
  Social Media Mining for Health Applications Workshop {\&} Shared Task}, pages
  9--12, Brussels, Belgium. Association for Computational Linguistics.

\bibitem[{Ramachandran et~al.(2023)Ramachandran, Lybarger, Liu, Mahajan, Liang,
  Tsou, Yetisgen, and Özlem Uzuner}]{RAMACHANDRAN2023104302}
Giridhar~Kaushik Ramachandran, Kevin Lybarger, Yaya Liu, Diwakar Mahajan,
  Jennifer~J. Liang, Ching-Huei Tsou, Meliha Yetisgen, and Özlem Uzuner. 2023.
\newblock \href {https://doi.org/https://doi.org/10.1016/j.jbi.2023.104302}
  {Extracting medication changes in clinical narratives using pre-trained
  language models}.
\newblock \emph{Journal of Biomedical Informatics}, 139:104302.

\bibitem[{Romberg et~al.(2020)Romberg, Dyczmons, Borgmann, Sommer, Vomhof,
  Brunoni, Bruck-Ramisch, Enders, Icks, and Conrad}]{romberg-etal-2020}
Julia Romberg, Jan Dyczmons, Sandra~Olivia Borgmann, Jana Sommer, Markus
  Vomhof, Cecilia Brunoni, Ismael Bruck-Ramisch, Luis Enders, Andrea Icks, and
  Stefan Conrad. 2020.
\newblock \href {https://aclanthology.org/2020.smm4h-1.3} {Annotating patient
  information needs in online diabetes forums}.
\newblock In \emph{Proceedings of the Fifth Social Media Mining for Health
  Applications Workshop {\&} Shared Task}, pages 19--26, Barcelona, Spain
  (Online). Association for Computational Linguistics.

\bibitem[{Roy and Pan(2021)}]{roy-pan-2021-incorporating}
Arpita Roy and Shimei Pan. 2021.
\newblock \href {https://doi.org/10.18653/v1/2021.emnlp-main.435}
  {Incorporating medical knowledge in {BERT} for clinical relation extraction}.
\newblock In \emph{Proceedings of the 2021 Conference on Empirical Methods in
  Natural Language Processing}, pages 5357--5366, Online and Punta Cana,
  Dominican Republic. Association for Computational Linguistics.

\bibitem[{Sampathkumar et~al.(2014)Sampathkumar, Chen, and
  Luo}]{sampathkumar2014mining}
Hariprasad Sampathkumar, Xue-wen Chen, and Bo~Luo. 2014.
\newblock \href {https://pubmed.ncbi.nlm.nih.gov/25341686/} {Mining adverse
  drug reactions from online healthcare forums using hidden markov model}.
\newblock \emph{BMC medical informatics and decision making}, 14(1):1--18.

\bibitem[{Shen and Rudzicz(2017)}]{shen-rudzicz-2017-detecting}
Judy~Hanwen Shen and Frank Rudzicz. 2017.
\newblock \href {https://doi.org/10.18653/v1/W17-3107} {Detecting anxiety
  through {R}eddit}.
\newblock In \emph{Proceedings of the Fourth Workshop on Computational
  Linguistics and Clinical Psychology {---} From Linguistic Signal to Clinical
  Reality}, pages 58--65, Vancouver, BC. Association for Computational
  Linguistics.

\bibitem[{Uzuner et~al.(2010)Uzuner, Solti, Xia, and Cadag}]{2009i2b2}
{\"O}zlem Uzuner, Imre Solti, Fei Xia, and Eithon Cadag. 2010.
\newblock \href {https://doi.org/10.1136/jamia.2010.004200} {Community
  annotation experiment for ground truth generation for the i2b2 medication
  challenge}.
\newblock \emph{J Am Med Inform Assoc JAMIA}, 17:519--23.

\bibitem[{Wadhwa et~al.(2022)Wadhwa, Khetan, Amir, and Wallace}]{redhot2022}
Somin Wadhwa, Vivek Khetan, Silvio Amir, and Byron Wallace. 2022.
\newblock \href {https://doi.org/10.48550/ARXIV.2210.06331} {Redhot: A corpus
  of annotated medical questions, experiences, and claims on social media}.

\bibitem[{Wolf et~al.(2020)Wolf, Debut, Sanh, Chaumond, Delangue, Moi
  et~al.}]{wolf-etal-2020-transformers}
Thomas Wolf, Lysandre Debut, Victor Sanh, Julien Chaumond, Clement Delangue,
  Anthony Moi, et~al. 2020.
\newblock \href {https://www.aclweb.org/anthology/2020.emnlp-demos.6}
  {Transformers: State-of-the-art natural language processing}.
\newblock In \emph{Conference on Empirical Methods in Natural Language
  Processing: System Demonstrations}, pages 38--45, Online. Association for
  Computational Linguistics.

\bibitem[{Zhong and Chen(2021)}]{zhong2021frustratingly}
Zexuan Zhong and Danqi Chen. 2021.
\newblock \href {https://doi.org/10.18653/v1/2021.naacl-main.5} {A
  frustratingly easy approach for entity and relation extraction}.
\newblock In \emph{Conference of the North American Chapter of the Association
  for Computational Linguistics: Human Language Technologies}, pages 50--61,
  Online. Association for Computational Linguistics.

\bibitem[{Zhou et~al.(2022)Zhou, Lybarger, Yetisgen, and
  Ostendorf}]{domain_generalization2022}
Sitong Zhou, Kevin Lybarger, Meliha Yetisgen, and Mari Ostendorf. 2022.
\newblock \href {https://arxiv.org/abs/2209.09485} {Generalizing through
  forgetting -- domain generalization for symptom event extraction in clinical
  notes}.

\bibitem[{Zirikly et~al.(2019)Zirikly, Resnik, Uzuner, and
  Hollingshead}]{zirikly-etal-2019-clpsych}
Ayah Zirikly, Philip Resnik, {\"O}zlem Uzuner, and Kristy Hollingshead. 2019.
\newblock \href {https://doi.org/10.18653/v1/W19-3003} {{CLP}sych 2019 shared
  task: Predicting the degree of suicide risk in {R}eddit posts}.
\newblock In \emph{Proceedings of the Sixth Workshop on Computational
  Linguistics and Clinical Psychology}, pages 24--33, Minneapolis, Minnesota.
  Association for Computational Linguistics.

\bibitem[{Zomick et~al.(2019)Zomick, Levitan, and
  Serper}]{zomick-etal-2019-linguistic}
Jonathan Zomick, Sarah~Ita Levitan, and Mark Serper. 2019.
\newblock \href {https://doi.org/10.18653/v1/W19-3009} {Linguistic analysis of
  schizophrenia in {R}eddit posts}.
\newblock In \emph{Proceedings of the Sixth Workshop on Computational
  Linguistics and Clinical Psychology}, pages 74--83, Minneapolis, Minnesota.
  Association for Computational Linguistics.

\end{thebibliography}
\bibliographystyle{acl_natbib}
\appendix

\end{document}